\documentclass{article}

\usepackage{arxiv}

\usepackage[utf8]{inputenc} 
\usepackage[T1]{fontenc}    
\usepackage{hyperref}       
\usepackage{url}            
\usepackage{booktabs}       
\usepackage{amsfonts}       
\usepackage{nicefrac}       
\usepackage{microtype}      
\usepackage{lipsum}
\usepackage{graphicx}
\usepackage{amsmath}
\graphicspath{ {./images/} }
\usepackage{tikz}
\usepackage{xcolor}
\usepackage{float}
\usepackage{tabularx} 
\usepackage{enumitem}
\usepackage{array}
\usepackage{makecell}
\usepackage{caption}
\usepackage{wrapfig}

\title{REGION IN CONTEXT: TEXT-CONDITIONED IMAGE EDITING WITH HUMAN-LIKE SEMANTIC REASONING}

\author{
Thuy Phuong Vu \\
Faculty of Electrical and Electronic Engineering\\
School of Engineering\\
Phenikaa University\\
Hanoi, Vietnam \\
\And
Dinh-Cuong Hoang \\
Greenwich Vietnam\\
FPT University\\
Hanoi, Vietnam \\
\And
Minhhuy Le\textsuperscript{*} \\
Faculty of Electrical and Electronic Engineering\\
School of Engineering\\
Phenikaa University\\
Hanoi, Vietnam \\
\texttt{huy.leminh@phenikaa-uni.edu.vn} \\
\And
Phan Xuan Tan\textsuperscript{*}\\
College of Engineering\\
Shibaura Institute of Technology\\
Tokyo, Japan\\
\texttt{tanpx@shibaura-it.ac.jp}
}

\begin{document}
\maketitle
\begingroup
\renewcommand\thefootnote{}\footnotetext{\textsuperscript{*}Corresponding authors.}
\endgroup
\begin{abstract}
Recent research has made significant progress in localizing and editing image regions based on text. However, most approaches treat these regions in isolation, relying solely on local cues without accounting for how each part contributes to the overall visual and semantic composition. This often results in inconsistent edits, unnatural transitions, or loss of coherence across the image. 
In this work, we propose Region in Context, a novel framework for text-conditioned image editing that performs multilevel semantic alignment between vision and language, inspired by the human ability to reason about edits in relation to the whole scene. Our method encourages each region to understand its role within the global image context, enabling precise and harmonized changes.
At its core, the framework introduces a dual-level guidance mechanism: regions are represented with full-image context and aligned with detailed region-level descriptions, while the entire image is simultaneously matched to a comprehensive scene-level description generated by a large vision-language model. These descriptions serve as explicit verbal references of the intended content, guiding both local modifications and global structure.
Experiments show that it produces more coherent and instruction-aligned results.
\noindent Code is available at: \href{https://github.com/thuyvuphuong/Region-in-Context.git}{https://github.com/thuyvuphuong/Region-in-Context.git}
\end{abstract}

\section{Introduction}
\label{sec:intro}
\begin{figure}[h]
    \centering
    \begin{minipage}{0.48\linewidth}
        \centering
        \includegraphics[width=\linewidth]{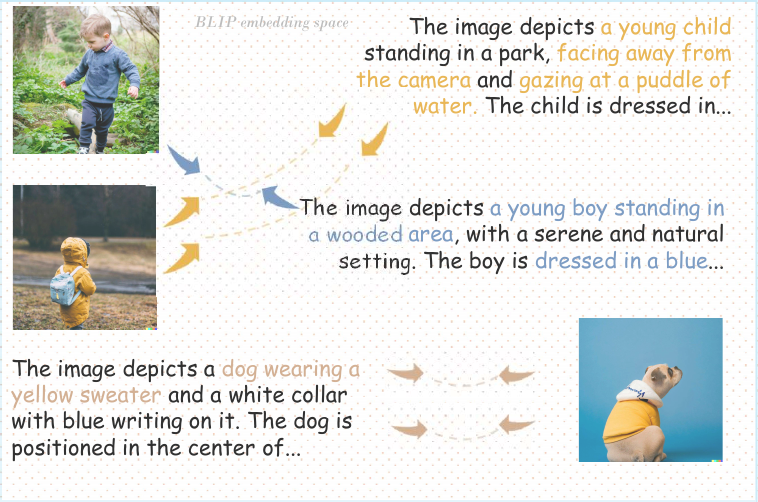}
    \end{minipage}%
    \hfill
    \begin{minipage}{0.52\linewidth}
        \centering
        \includegraphics[width=\linewidth]{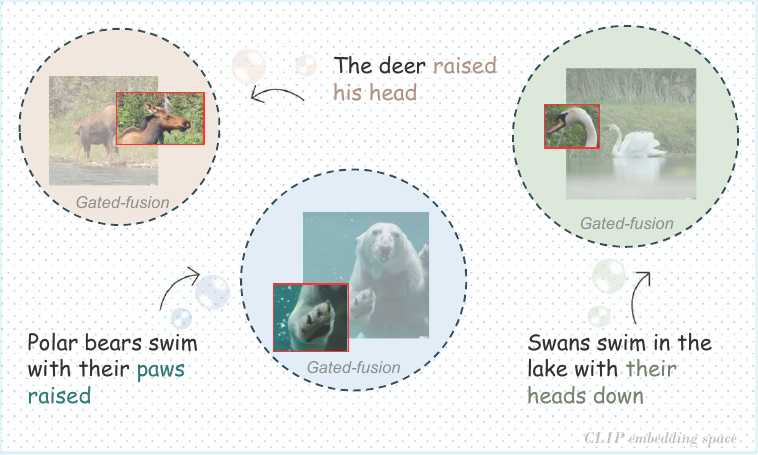}
    \end{minipage}
    \caption{Illustration of our vision-language embedding optimization strategy. At the global level (left), the model aligns image and full target description embeddings. At the local level (right), it aligns the region in the full image to make the region \textit{"see itself in context"}, with their corresponding region descriptions using contrastive learning.}
    \label{fig:embedding_optimize}
\end{figure}
Just as solving a puzzle requires understanding the context of the entire image to determine where each piece belongs, many human visual tasks involve reasoning about the whole before acting on the parts. We do not place a puzzle piece based solely on its shape or color in isolation; instead, we look at the full picture to infer its correct position. This process reflects a broader principle in visual understanding: local decisions are guided by global context. Whether assembling a puzzle or arranging elements in a design, humans rely on a mental model of the complete scene to ensure that each part contributes meaningfully and harmoniously to the whole. This same principle applies to the task of image editing. When modifying a specific region within an image—such as changing the color of an object or altering its appearance—humans rarely consider that region in isolation. Instead, we take into account the broader scene to ensure that the edit fits naturally, both visually and semantically, within the overall composition as Fig. \ref{fig:embedding_optimize}.\\

\begin{wrapfigure}{l}{0.5\linewidth}
    \centering
    \begin{minipage}[c]{0.3\linewidth}
    \centering
    \includegraphics[width=\linewidth]{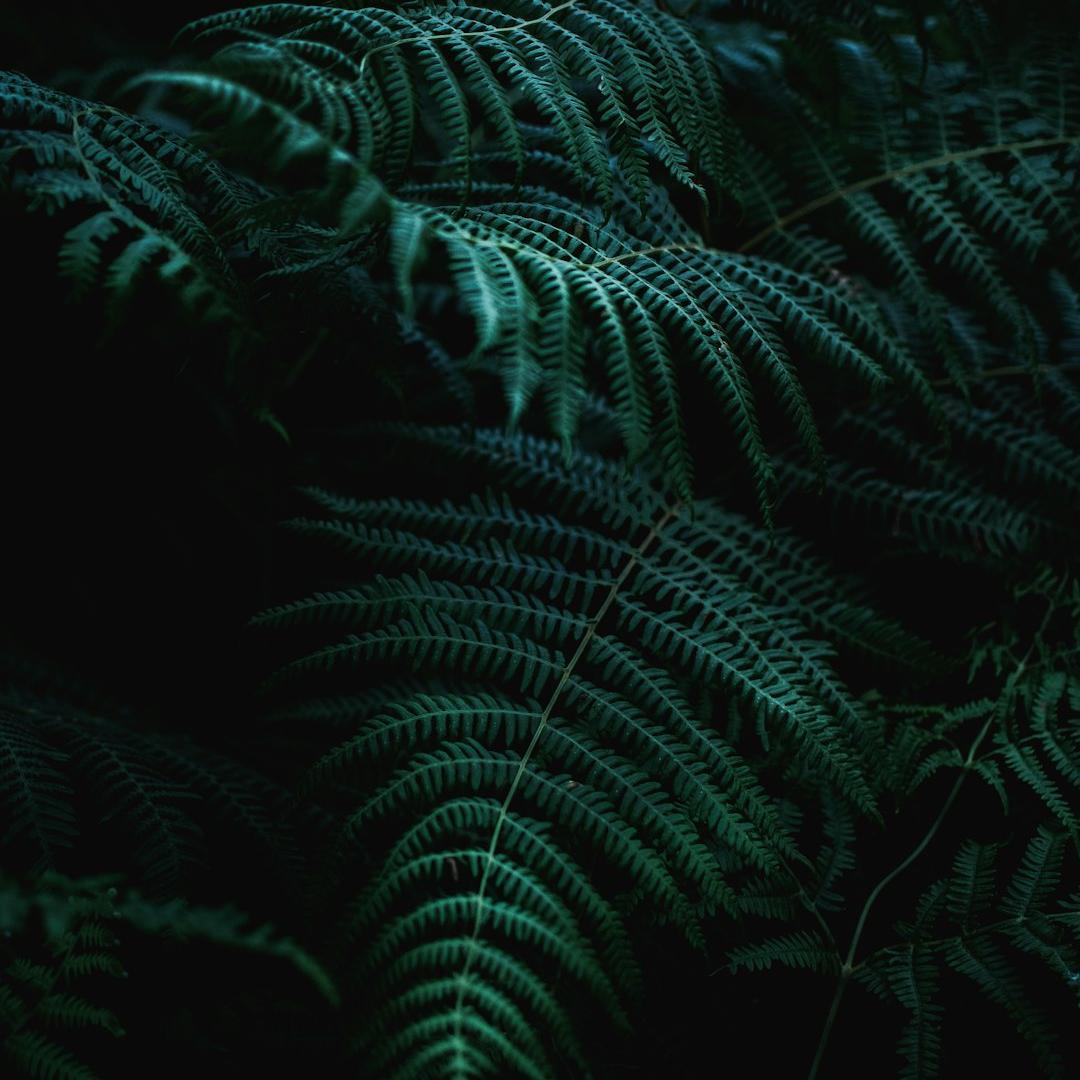}
    \caption*{Add\\ \textcolor{red}{butterflies}}
    \end{minipage}
    \hfill
    \begin{minipage}[c]{0.3\linewidth}
        \centering
        \includegraphics[width=\linewidth]{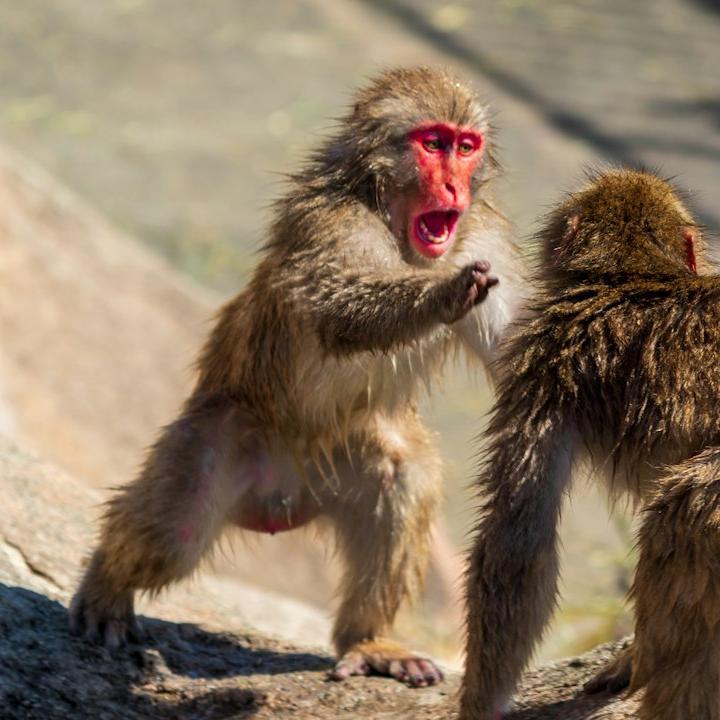}
        \caption*{\centering Close \textcolor{red}{monkey} \textcolor{red}{mouth}}
    \end{minipage}
    \hfill
    \begin{minipage}[c]{0.3\linewidth}
        \centering
        \includegraphics[width=\linewidth]{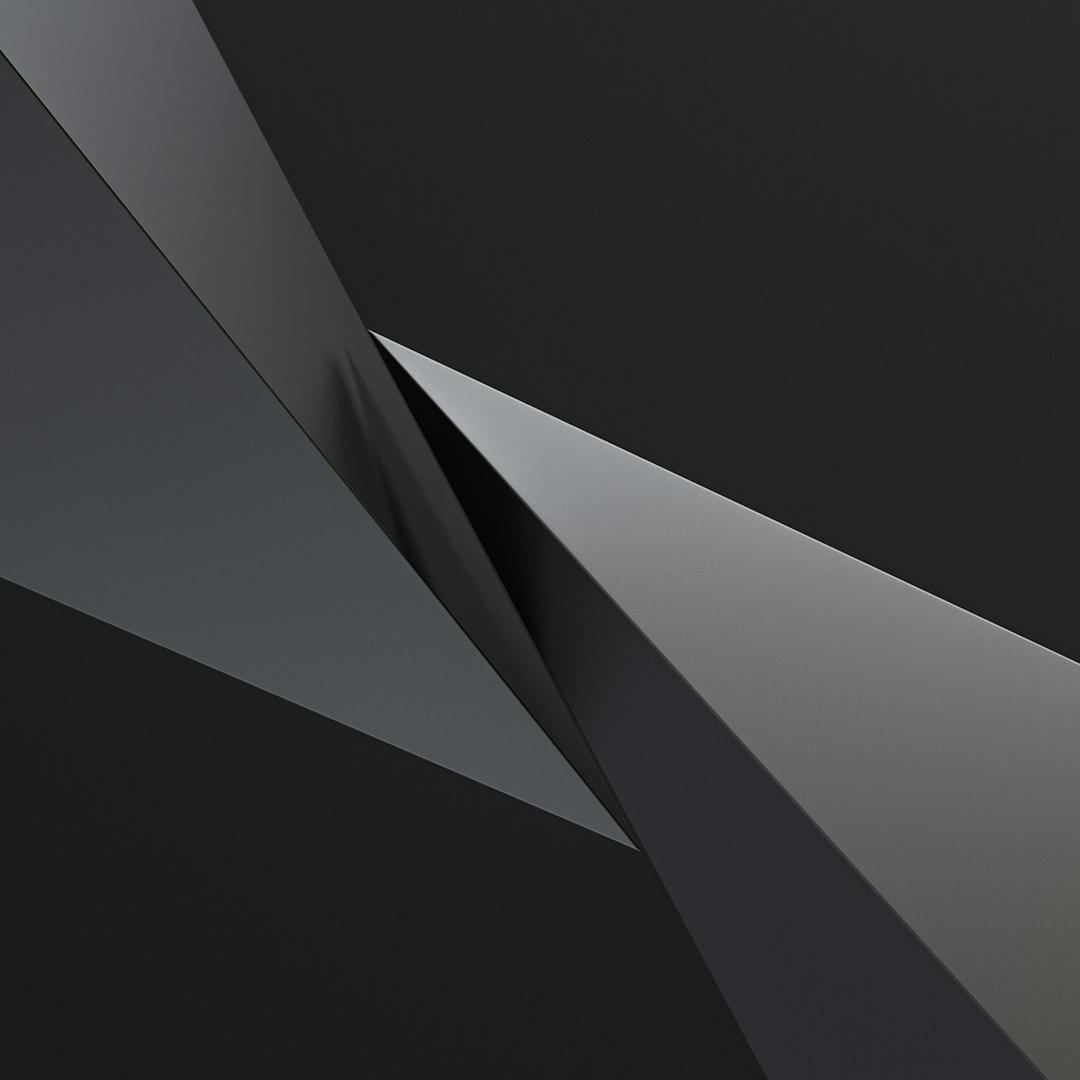}
        \caption*{\centering Introduce \\ \textcolor{red}{insects}}
    \end{minipage}

    \vspace{1ex}
    \begin{tikzpicture}

        \draw[->, thick] (1, 0) -- (1, -0.3);
        \draw[->, thick] (4, 0) -- (4, -0.3);
        \draw[->, thick] (7, 0) -- (7, -0.3);
    \end{tikzpicture}
    \vspace{1ex}

    \begin{minipage}[c]{0.3\linewidth}
        \centering
        \includegraphics[width=\linewidth]{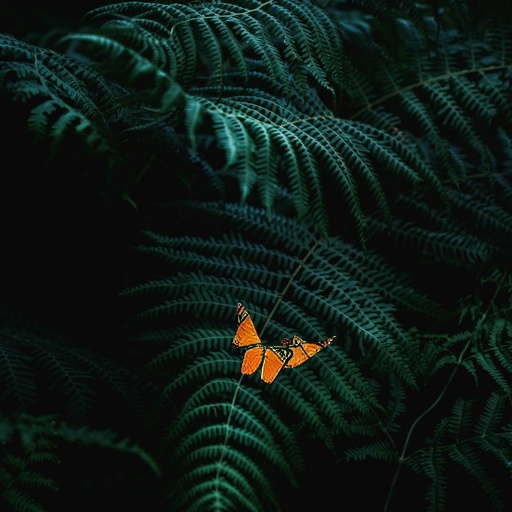}
    \end{minipage}
    \hfill
    \begin{minipage}[c]{0.3\linewidth}
        \centering
        \includegraphics[width=\linewidth]{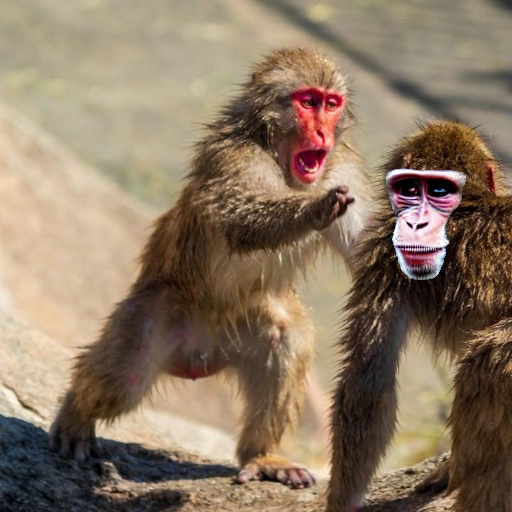}
    \end{minipage}
    \hfill
    \begin{minipage}[c]{0.3\linewidth}
        \centering
        \includegraphics[width=\linewidth]{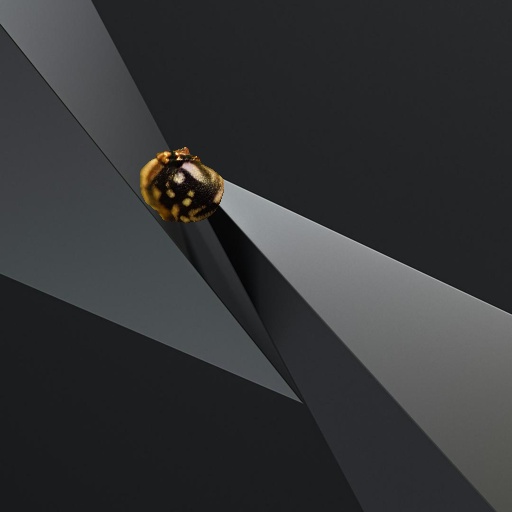}
    \end{minipage}

    \caption{Illustration of editing results using the segmentation mask from ZONE \cite{ZONE}, which first segments the editing region, then edits and blends it in the original image. Although the method includes a blending step during post-processing, the edited images still exhibit unnatural appearances.}
    \vspace{-0.5cm}

    \label{fig:problem}
\end{wrapfigure}

However, many recent methods for text-conditioned image editing focus primarily on localizing and modifying target regions \cite{Liu_2024_CVPR, ZONE}, often relying on isolated region embeddings or mask-based attention. While these approaches enable targeted control, they frequently lack mechanisms to reason about the role of the region within the full image. As shown in Fig. \ref{fig:problem}, even with post-processing techniques such as blending~\cite{ZONE}, the result may appear unnatural, like a sticker pasted onto the image. Other works, such as \cite{foi, pair}, still fall short in achieving scene-level semantic coherence. Moreover, approaches like \cite{text-driven} focus on learning localized attention masks, yet remain limited in modeling relationships between the edited region and the overall visual semantics.

Vision and language are fundamentally intertwined—humans often interpret visual scenes through verbal descriptions and, conversely, imagine visuals based on textual cues \cite{clip, li2022blip, li2023blip, alayrac2022flamingo, lu2024deepseek}. This close bond has inspired growing interest in integrating language understanding into visual generation systems, particularly for tasks like text-conditioned image editing. \cite{fu2024guidinginstructionbasedimageediting, SLD}. Broadly, existing approaches integrate LLMs in two main ways. First, several works use LLMs to improve dataset construction by refining or generating detailed editing instructions and compositional prompts \cite{ip2p, bai2025humanedithighqualityhumanrewardeddataset, fu2024guidinginstructionbasedimageediting}. Second, a growing line of research embeds LLMs directly within the editing pipeline to act as reasoning agents or feedback mechanisms. Notably, SLD \cite{SLD} introduces an iterative feedback loop where the LLM evaluates the consistency of the generated image with the input prompt and issues correctional instructions to the diffusion model. 

With these inspirations, in this paper, we propose a framework for image editing (Sec. \ref{framework}) that leverages language not only as an instruction medium but also as a semantic reference for global guidance. Specifically, we introduce a dual-level alignment mechanism that encourages each edited region to be understood in the context of the full image by introducing the gated region-context fusion module (Sec. \ref{regionincontext}), while also aligning the entire edited result with a scene-level description (Sec. \ref{embedding}) by optimizing the vision-language embedding models, i.e., CLIP \cite{clip} and BLIP \cite{li2022blip}, with contrastive learning. These verbal references are automatically generated using large language models, i.e., Deepseek VL \cite{lu2024deepseek}, based on structured prompts as described in Sec. \ref{llm} and serve as grounding anchors during the editing process. By comparing the region-in-context with its corresponding region-level description and the full image with the global description, our approach ensures that edits are both precise and semantically coherent. This design reflects how humans evaluate edits—not in isolation, but through an understanding of the broader visual and linguistic context.

\section{Related works}
\label{sec:relatedworks}

Text-conditioned image editing with diffusion models has evolved through various strategies to balance edit quality, controllability, and user intent. Sampling-time or inversion-based methods, such as~\cite{huberman2024edit, lee2025diffusion, foi, brack2024leditslimitlessimageediting, chen2024tino}, avoid costly retraining by manipulating embeddings, noise schedules, or guidance strength during inference. However, their effectiveness heavily relies on the pretrained noise prediction model, which can limit editing flexibility and fidelity in complex scenarios. Fine-tuning-based methods such as~\cite{ip2p, zhang2024magicbrushmanuallyannotateddataset, kawar2023imagic} enable instruction-following edits by training on paired image–instruction data, supporting local editing, user interactions, and multi-instruction control. These approaches are particularly effective, as they directly adapt the model’s behavior during training, allowing for purposeful improvements through new data or customized learning strategies. To support evaluation and training grounded in real user intent, HumanEdit \cite{bai2025humanedithighqualityhumanrewardeddataset} introduces a high-quality, human-rewarded dataset of image edits with natural instructions and diverse semantic transformations. Therefore, in this paper, we adopt this training approach to make the model reach our hypothesis.

A more specific purpose, such as region- and object-level editing, is addressed by models like \cite{ZONE, Liu_2024_CVPR, pair, text-driven}, which enhance spatial control through semantic decomposition or multimodal supervision. Meanwhile, methods such as \cite{Nam_2024_CVPR, hertz2022prompttopromptimageeditingcross, Liu_2024_CVPR} explore cross-attention modulation and denoising mechanisms to improve edit faithfulness and structural consistency.
More specifically, ZONE \cite{ZONE} proposes a zero-shot, segmentation-free approach that localizes instruction-relevant regions by leveraging CLIP-guided attention maps from a pretrained InstructPix2Pix \cite{ip2p} model. It further refines the region using Region-IoU and ensures smooth transitions via FFT-based edge blending, all without extra training or prompt tuning. While \cite{text-driven, pair} attempt to enhance regional focus through attention maps or using multimodal to extract more information to add to the model as an additional condition, they do not explicitly supervise whether the model has effectively learned to edit within the intended area. A more refined strategy is proposed in \cite{Liu_2024_CVPR}, where regional conditions are injected into the noise prediction model and guided through a self-supervised training pipeline. However, this supervision is applied in isolation, using only binary masks to evaluate whether the edit occurred in the correct region—without considering how that region interacts with the overall image context.

To address this limitation, we propose a training framework that encourages the model to reason about each editable region in the context of the entire image, aligning it not only with spatial constraints but also with the intended semantic meaning of both the region and the global scene.

\section{Proposals}
\subsection{Preliminaries}
Diffusion models generate data by simulating a Markov chain of latent variables $\mathbf{x}_1, \dots, \mathbf{x}_T$ starting from clean input and progressively adding Gaussian noise at each time step $t$, as in Eq. \ref{eq:diff1}, where $\beta_t$ is a small positive variance schedule. The full forward process is defined as Eq. \ref{eq:diff2} with $q(\mathbf{x}_0)$ being the real data distribution. As shown in Eq. \ref{eq:diff_reverse}, the reverse process learns to denoise $\mathbf{x}_t$ step-by-step back to $\mathbf{x}_0$ using a neural network $\epsilon_\theta$ to predict the noise, where $\bar{\alpha}_t = \prod_{s=1}^{t}(1 - \beta_s)$ and $\mathbf{z} \sim \mathcal{N}(\mathbf{0}, \mathbf{I})$.
\begin{equation}
q(\mathbf{x}_t \mid \mathbf{x}_{t-1}) = \mathcal{N}(\mathbf{x}_t; \sqrt{1 - \beta_t} \, \mathbf{x}_{t-1}, \beta_t \mathbf{I}),
\label{eq:diff1}
\end{equation}
\begin{equation}
q(\mathbf{x}_{1:T} \mid \mathbf{x}_0) = \prod_{t=1}^T q(\mathbf{x}_t \mid \mathbf{x}_{t-1}),
\label{eq:diff2}
\end{equation}
\begin{equation}
\mathbf{x}_{t-1} = \frac{1}{\sqrt{1 - \beta_t}} \left( \mathbf{x}_t - \frac{\beta_t}{\sqrt{1 - \bar{\alpha}_t}} \epsilon_\theta(\mathbf{x}_t, t) \right) + \sigma_t \mathbf{z},
\label{eq:diff_reverse}
\end{equation}
\paragraph{Text-Conditioned Image Editing.} To enable text-conditioned generation or editing, the noise prediction network $\epsilon_\theta$ is conditioned on a text prompt embedding $\mathbf{e}_\text{prompt}$ as Eq. \ref{eq:ie}. In image editing, the input is typically an image $\mathbf{x}$ and an instruction $\mathbf{e}_\text{text}$ that specifies the desired modification. One common approach is to encode $\mathbf{x}$ into a latent representation $\mathbf{z}_T$ through a forward diffusion process and then perform denoising conditioned on $\mathbf{y}$ to obtain the edited image $\hat{\mathbf{x}}_0$.
\begin{equation}
\epsilon_\theta(\mathbf{x}_t, t, \mathbf{e}_\text{prompt}) \approx \epsilon.
\label{eq:ie}
\end{equation}
\subsection{Our framework}
\begin{figure*}[t]
    \centering
    \includegraphics[width=\linewidth]{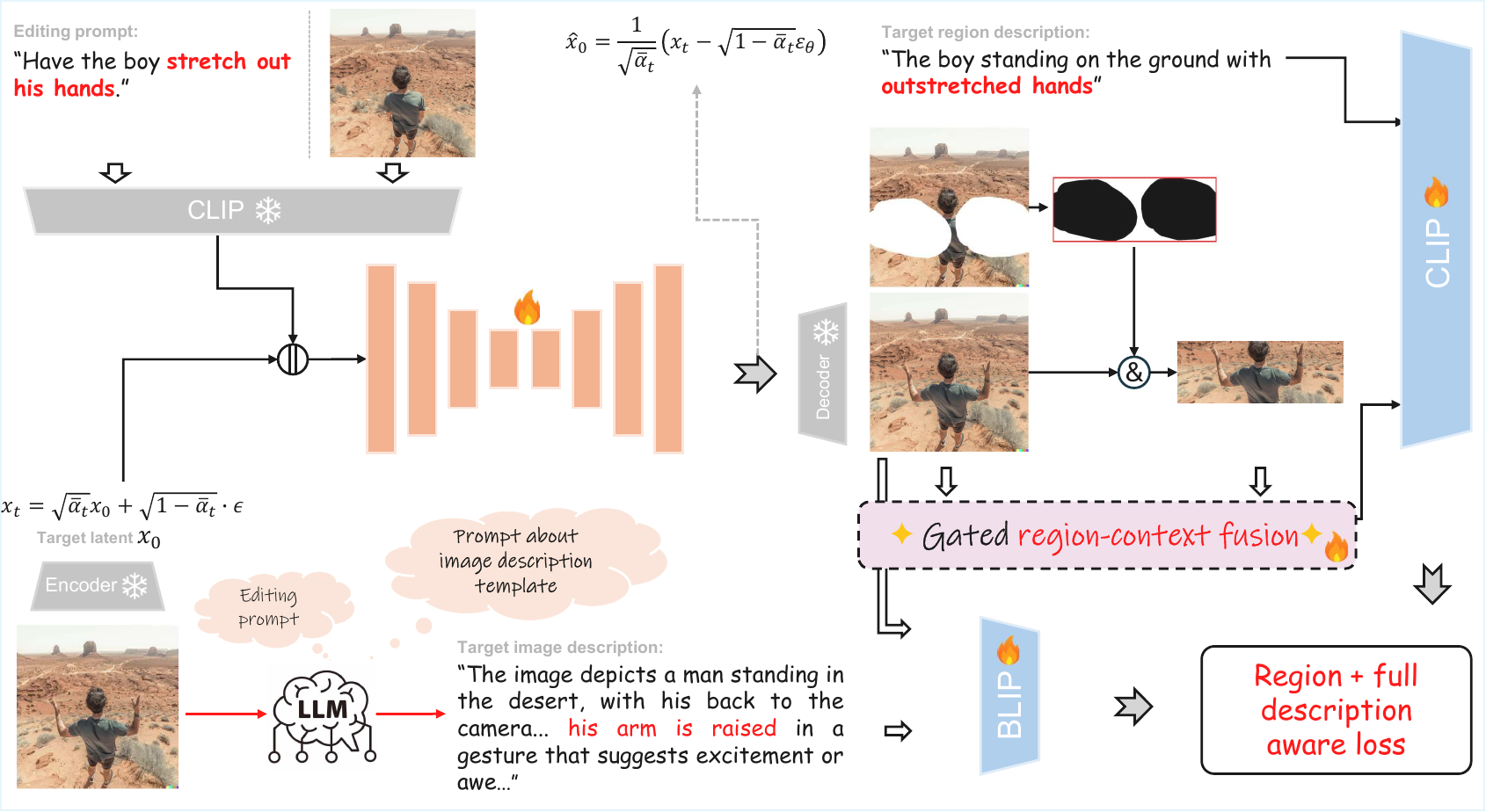}
    \caption{The proposed region-aware diffusion framework for text-guided image editing. The model aligns visual and textual representations at both region and scene levels using a gated region-context fusion module and contrastive supervision. A denoising process reconstructs the edited image from noisy input while ensuring semantic fidelity through region-level and global alignment losses.}
    \label{fig:model_architecture}
    \vspace{-0.3cm}
\end{figure*}
We propose a region-aware diffusion model training framework shown as Fig. \ref{fig:model_architecture} for text-guided image editing that jointly aligns image and text representations at both global and regional levels. Given a noisy latent input \( x_t \), our model reconstructs the clean latent \( \hat{x}_0 \) using the denoising equation as Eq. \ref{eq:denoise}, where \( \epsilon_\theta \) is the noise prediction network conditioned on the text prompt \( c \), and \( \bar{\alpha}_t \) is the cumulative product of the noise schedule up to timestep \( t \). 
\begin{equation}
    \hat{x}_0 = \frac{1}{\sqrt{\bar{\alpha}_t}} \left( x_t - \sqrt{1 - \bar{\alpha}_t} \cdot \epsilon_\theta(x_t, t, c) \right),
    \label{eq:denoise}
\end{equation}
To capture fine-grained semantics, both the full image description and the region-specific instruction are embedded and fused using a gated region-context mechanism. Let the predicted clean latent be \( \hat{x}_0 \in \mathbb{R}^{H \times W \times C} \), and a bounding box defined as \( \mathbf{b} = (x_{\text{min}}, y_{\text{min}}, x_{\text{max}}, y_{\text{max}}) \), which is computed as the tightest box enclosing the region mask \( M \in \{0, 1\}^{H \times W} \). The cropped region corresponding to \( \mathbf{b} \) is given by Eq.~\ref{eq:bbox}. The gated-fused region is computed in Eq.~\ref{eq:fuse}, where \( F \) denotes the gated region-context fusion module. To supervise region-level alignment, we compute the distance between the gated-fused region embedding \( f_r \) and the embedding of the target region description \( \mathbf{e}_{t_r} \), encoded via an \textbf{optimized} shared CLIP embedding space encoder in Sec. \ref{embedding}. The region loss is defined as the distance between these two embeddings in Eq.~\ref{eq:region_loss}. The global loss is the distance on the \textbf{optimized} BLIP embedding space between the full image $\hat{x}_0$ and its corresponding target description as Eq. \ref{eq:global_loss}, where $\mathbf{e}_{f}$ and $\mathbf{e}_{t_f}$ is the embedding of full image $\hat{x}_0$ and the target full image description $t_f$, respectively.
\begin{equation}
\hat{x}_0^{\mathbf{b}} = \hat{x}_0[y_{\text{min}} : y_{\text{max}}, \, x_{\text{min}} : x_{\text{max}}]
\label{eq:bbox}
\end{equation}
\begin{equation}
f_r = F(\hat{x}_0, \hat{x}_0^{\mathbf{b}})
\label{eq:fuse}
\end{equation}
\begin{equation}
\mathcal{L}_{\text{region}} = 1 - \cos(f_r, \mathbf{e}_{t_r}) = 1 - \frac{f_r^\top \mathbf{e}_{t_r}}{\|f_r\|_2 \cdot \|\mathbf{e}_{t_r}\|_2}
\label{eq:region_loss}
\end{equation}
\begin{equation}
\mathcal{L}_{\text{global}} = 1 - \cos(\mathbf{e}_{f}, \mathbf{e}_{t_f}) = 1 - \frac{\mathbf{e}_{f}^\top \mathbf{e}_{t_f}}{\|\mathbf{e}_{f}\|_2 \cdot \|\mathbf{e}_{t_f}\|_2}
\label{eq:global_loss}
\end{equation}
The model is trained using a combination of region-aware loss, full description alignment loss, and denoising loss illustrated in Eq. \ref{eq:total_loss}. These objectives jointly minimize the distance between matching image-text pairs while pushing apart mismatched ones in the embedding space, and ensure consistency between the added noise and the predicted noise during the diffusion process.
\begin{equation}
\begin{aligned}
\mathcal{L}_{\text{total}} = \; & 
\mathcal{L}_\text{region} + \mathcal{L}_\text{global}
+ \; \mathbb{E}_{x_0, t, \epsilon} \left[ \left\| \epsilon_\theta(x_t, t, c) - \epsilon \right\|_2^2 \right]
\end{aligned}
\label{eq:total_loss}
\end{equation}
\label{framework}
\subsubsection{Vision-language embedding optimization}
\label{embedding}
To promote semantic consistency between the generated visual content and its intended meaning, we optimize a vision-language embedding space where both image regions and scene-level representations are aligned with their corresponding target textual descriptions. This optimization is carried out independently for local (region-level) and global (scene-level) alignment using pretrained multimodal encoders.

For region-level alignment, we embed context-aware image patches $f_r$ and their respective target region description $t_r$ into a shared space using a modified CLIP-based model with gated cross-attention fusion. Similarly, for scene-level alignment, we use a pretrained BLIP model to embed the entire image $\hat{x}_0$ and a detailed scene description $t_f$. While CLIP is effective for learning joint embeddings of image-text pairs, it has a limitation in the number of input tokens it can process (77 tokens), which constrains the level of detail that can be captured in textual descriptions. This makes CLIP more suitable for short, focused region descriptions. In contrast, BLIP is designed to handle longer and more descriptive text inputs (512 tokens), making it better suited for scene-level alignment, where a more comprehensive understanding of the global image context and detailed descriptions is required. In both cases, alignment is enforced using a symmetric contrastive loss defined as Eq. \ref{eq:contrastive_loss}, encouraging the embeddings of matching image–text pairs to be closer than those of mismatched ones. 
\begin{equation}
\begin{aligned}
\mathcal{L}_{\text{optim}} = \frac{1}{2N} \sum_{i=1}^N \Big[ 
& - \log \frac{\exp\left( \mathbf{z}_i^\top \mathbf{t}_i / \tau \right)}{\sum_{j=1}^N \exp\left( \mathbf{z}_i^\top \mathbf{t}_j / \tau \right)} \\
& - \log \frac{\exp\left( \mathbf{t}_i^\top \mathbf{z}_i / \tau \right)}{\sum_{j=1}^N \exp\left( \mathbf{t}_i^\top \mathbf{z}_j / \tau \right)} \Big]
\end{aligned}
\label{eq:contrastive_loss}
\end{equation}
\begin{wrapfigure}{r}{0.5\linewidth}
    \centering
    \includegraphics[width=0.95\linewidth]{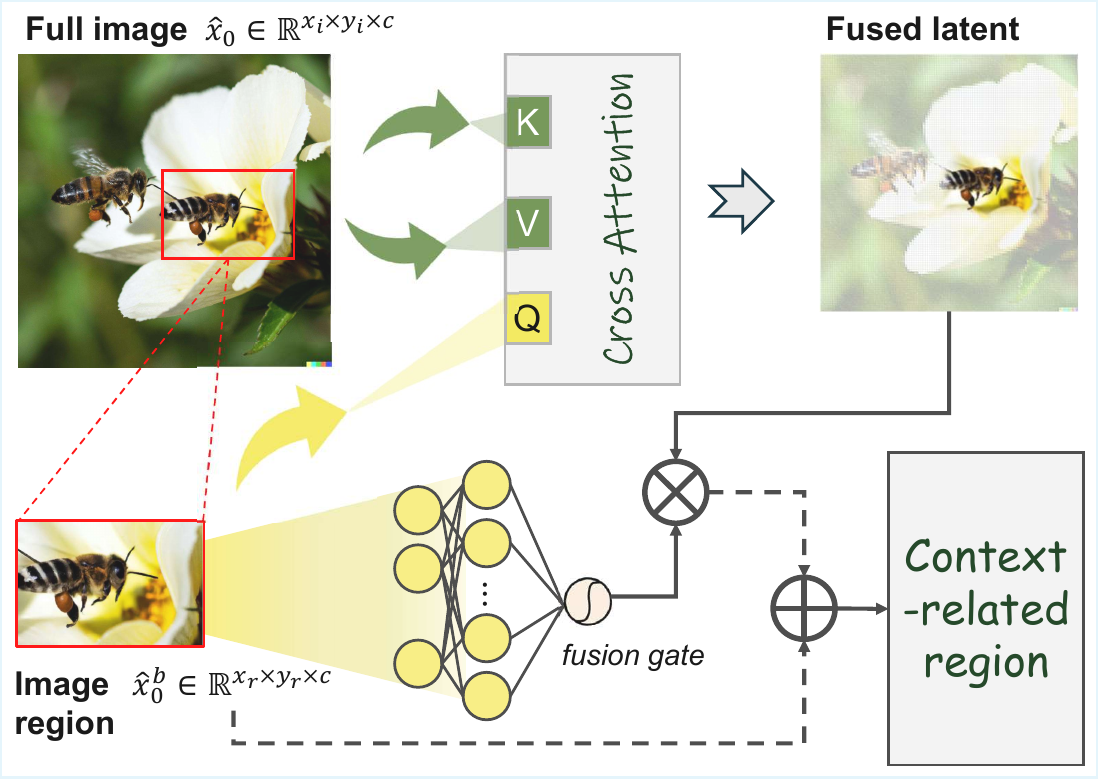}
    \caption{Gated Cross-Attention Fusion module. Region and full image embeddings are first transformed, then fused via multi-head cross-attention. A learnable gate regulates the influence of global context on the region, ensuring controlled and context-aware representation refinement.}
    \label{fig:gated_fusion}
    \vspace{-1cm}
\end{wrapfigure}

Generally, $\mathbf{z}_i$ denotes the visual embedding for the $i$-th sample. The contrastive loss over a batch of $N$ image–text pairs, where $\tau$ is a temperature parameter that scales the similarity scores. This formulation is applied independently to both region-level embeddings (using CLIP + fusion) and scene-level embeddings (using BLIP), ensuring that the model aligns visual and textual semantics at multiple levels of granularity.

\subsubsection{Gated region in context}
\label{regionincontext}
With the optimized embedding model with the vision-language embedding $\mathbf{e}$, to incorporate global context into localized region representations while maintaining semantic focus, we propose a \textit{Gated Cross-Attention Fusion} module as shown in Fig. \ref{fig:gated_fusion}. To allow the feature of region $\mathbf{e}_r$ to get understand its role within the broader scene $\mathbf{e}_f$, we model its feature $\tilde{\mathbf{e}}_f$ (through Eq. \ref{eq:cnns}) interaction with the full image using multi-head cross-attention. In this setup in Eq. \ref{eq:multihead}, the region embedding acts as the query, meaning it asks: \textit{''Which part of the full image is relevant to me?''} The full image embedding provides both the keys and values, representing the available contextual information. Through this mechanism, the region learns to attend to the most semantically relevant aspects of the full image, enabling it to refine its representation based on the scene in which it appears.
\begin{equation}
\tilde{\mathbf{e}}_r = \mathrm{CNN}_r(\mathbf{e}_r), \quad \tilde{\mathbf{e}}_f = \mathrm{CNN}_f(\mathbf{e}_f)
\label{eq:cnns}
\end{equation}
\begin{equation}
\mathbf{h} = \mathrm{MultiHeadAttn}(\text{Q = }\tilde{\mathbf{e}}_r,\ \text{K = }\tilde{\mathbf{e}}_f,\ \text{V = }\tilde{\mathbf{e}}_f)
\label{eq:multihead}
\end{equation}
To avoid over-reliance on contextual signals, we introduce a gating mechanism as Eq. \ref{eq:gate} that controls the influence of the attended global information. This formulation through Eq. \ref{eq:gated} and \ref{eq:gated2} ensures that the region embedding is selectively enhanced by global image context, enabling coherent and context-aware modifications without compromising local semantic fidelity.
\begin{equation}
\mathbf{g} = \sigma(\mathbf{W}_g \tilde{\mathbf{e}}_r + \mathbf{b}_g)
\label{eq:gate}
\end{equation}
\begin{equation}
\mathbf{z} = \tilde{\mathbf{e}}_r + \mathbf{g} \odot \mathbf{h}
\label{eq:gated}
\end{equation}
\begin{equation}
f_r = \mathbf{W}_p \cdot \mathrm{LayerNorm}(\mathbf{z}) + \mathbf{b}_p
\label{eq:gated2}
\end{equation}

\subsubsection{Global LLM reference}
\label{llm}
To provide a rich semantic grounding for scene-level image alignment, we generate a detailed textual description of each image using a large vision-language model (VLM). Specifically, we employ DeepSeek-VL \cite{lu2024deepseek}, a multimodal causal language model, to produce structured paragraph-style captions that capture fine-grained visual content from the full image. Given an image and its editing instruction, we construct a carefully designed prompt as Table \ref{tab:llmprompt} that guides the model to describe observable attributes—such as objects, textures, colors, spatial relationships, and region-specific edits—without inferring unobservable factors. This process yields dense scene-level descriptions that serve as verbal references for aligning generated images during training. These LLM-generated descriptions are then embedded using a pretrained BLIP encoder and used in the global contrastive loss (Eq.~\ref{eq:global_loss}) to ensure that the semantic content of the generated image aligns with the intended overall meaning. This global reference complements the localized region-text alignment, allowing the model to reason about edits both precisely and contextually.
\subsection*{Implementation details}
\label{implement}
In the image description task, the model LLM is queried via a chat-style API using multimodal input, where the image and prompt are encoded with DeepSeek-VL's processor and passed through a pretrained DeepSeek-VL2-Tiny checkpoint \cite{wu2024deepseekvl2mixtureofexpertsvisionlanguagemodels}.

We train our Region-CLIP model using the HumanEdit dataset \cite{bai2025humanedithighqualityhumanrewardeddataset}, which provides paired images, target region-level instructions, and binary segmentation masks. All images and masks are resized to a resolution of $512 \times 512$. We use a pretrained CLIP ViT-B/16 model \cite{clip}. Training is conducted in two phases. In \textit{Phase 1}, the CLIP backbone is frozen and only the fusion module is optimized for 50 epochs using the AdamW optimizer with a learning rate of $1 \times 10^{-4}$ and a batch size of 64. In \textit{Phase 2}, both the CLIP encoder and the fusion module are fine-tuned jointly for 20 additional epochs with a learning rate of $1 \times 10^{-5}$. It takes 7.78 and 3.47 hours to train the fusion module and the whole process, respectively. With the global embedding model, BLIP (\texttt{blip-itm-base-coco}) \cite{li2022blip} is fine-tuned with a batch size of 32 and train for 20 epochs using the AdamW optimizer with a learning rate of $1 \times 10^{-5}$. The total time for training BLIP is 3.5 hours.

In training the noise prediction model, the pretrained UNet is initialized from a prior checkpoint and optimized using a learning rate of $1 \times 10^{-4}$ over 9000 steps. We use a resolution of $256 \times 256$, a batch size of 8, in approximately 16 training hours. The training loop uses the \texttt{accelerate} library \cite{accelerate} with XFormers' memory-efficient attention. All models were trained on a single NVIDIA A100 GPU.
\section{Evaluation}
\paragraph{Baselines and evaluation metrics.}
We evaluate our method against several strong text-conditioned image editing baselines, including InstructPix2Pix~\cite{ip2p}, MagicBrush~\cite{zhang2024magicbrushmanuallyannotateddataset}, ZONE~\cite{ZONE}, CDS~\cite{Nam_2024_CVPR}, and DDPM Inversion~\cite{huberman2024edit}. For fair comparison, we report results for our framework integrated into each baseline backbone, referred to as \textit{Ours (on IP2P)}, \textit{Ours (on MB)}, and \textit{Ours (on ZONE)}. Evaluation is conducted on the HumanEdit benchmark using a set of both language–image and perceptual quality metrics. We adopt six standard evaluation metrics: CLIP-I and CLIP-T measure CLIP-based similarity between the edited image and the input image or the target text, respectively, as in~\cite{clip}; DINO score~\cite{dino} assesses semantic consistency using DINO-ViT features; LPIPS~\cite{lpips} captures perceptual similarity; FID~\cite{fid} evaluates distributional realism against the ground-truth images; and IS~\cite{is} (Inception Score) quantifies the diversity and quality of generated outputs.
\begin{table*}[t]
\caption{Quantitative comparison of our proposal against other existing baselines. We evaluate performance across six metrics: CLIP-I, DINO, CLIP-T (semantic alignment), LPIPS (perceptual similarity), FID (distributional realism), and IS (image diversity and quality). Our training framework is applied to multiple pretrained diffusion backbones, showing consistent improvements in both semantic and perceptual metrics.}
\label{tab:quantitative_eval}
\centering
\begin{tabular}{lllllll}
\hline
\hline
\multicolumn{1}{c}{\textbf{Model}} & \multicolumn{1}{c}{\textbf{CLIP-I↑}} & \multicolumn{1}{c}{\textbf{DINO↑}} & \multicolumn{1}{c}{\textbf{CLIP-T↑}} & \multicolumn{1}{c}{\textbf{LPIPS↓}} & \multicolumn{1}{c}{\textbf{FID↓}} & \multicolumn{1}{c}{\textbf{IS↑}} \\ 
\hline
\makecell[l]{InstructPix2Pix \\ \cite{ip2p}} & 0.7531 & 0.8900 & 0.8390 & 0.5008 & 139.60 & 9.1981 \\
\textbf{Ours (on IP2P)} & \textbf{0.9146} & \textbf{0.9736} & \textbf{0.9337}  & \textbf{0.2042} & \textbf{77.07} & \textbf{10.4931} \\ \hline
\makecell[l]{MagicBrush \\ \cite{zhang2024magicbrushmanuallyannotateddataset}} & 0.8939 & 0.9646 & 0.9057 & 0.2448 & 88.96 & \textbf{9.8105} \\
\textbf{Ours (on MB)} & \textbf{0.9040} & \textbf{0.9664} & \textbf{0.9207} & \textbf{0.2263} & \textbf{85.61} & 9.6078 \\ \hline
ZONE \cite{ZONE} & 0.9569 & 0.9906 & 0.9319 & 0.0650 & 44.63 & 10.7947 \\
\textbf{Ours (on ZONE)} & \textbf{0.9729} & \textbf{0.9955} & \textbf{0.9332} & \textbf{0.0409} & \textbf{27.53} & \textbf{11.3038} \\ 
\hline
CDS \cite{Nam_2024_CVPR} & 0.9605 & 0.9941 & 0.9239 & 0.0938 & 29.50 & 10.9578\\
\makecell[l]{DDPM Inverse \\ \cite{huberman2024edit}} & 0.8433 & 0.9611 & 0.8584 & 0.2567 & 105.45 & 10.7182 \\
\hline
\hline
\end{tabular}
\end{table*}
\begin{figure*}[h]
    \centering
    \includegraphics[width=0.9\linewidth]{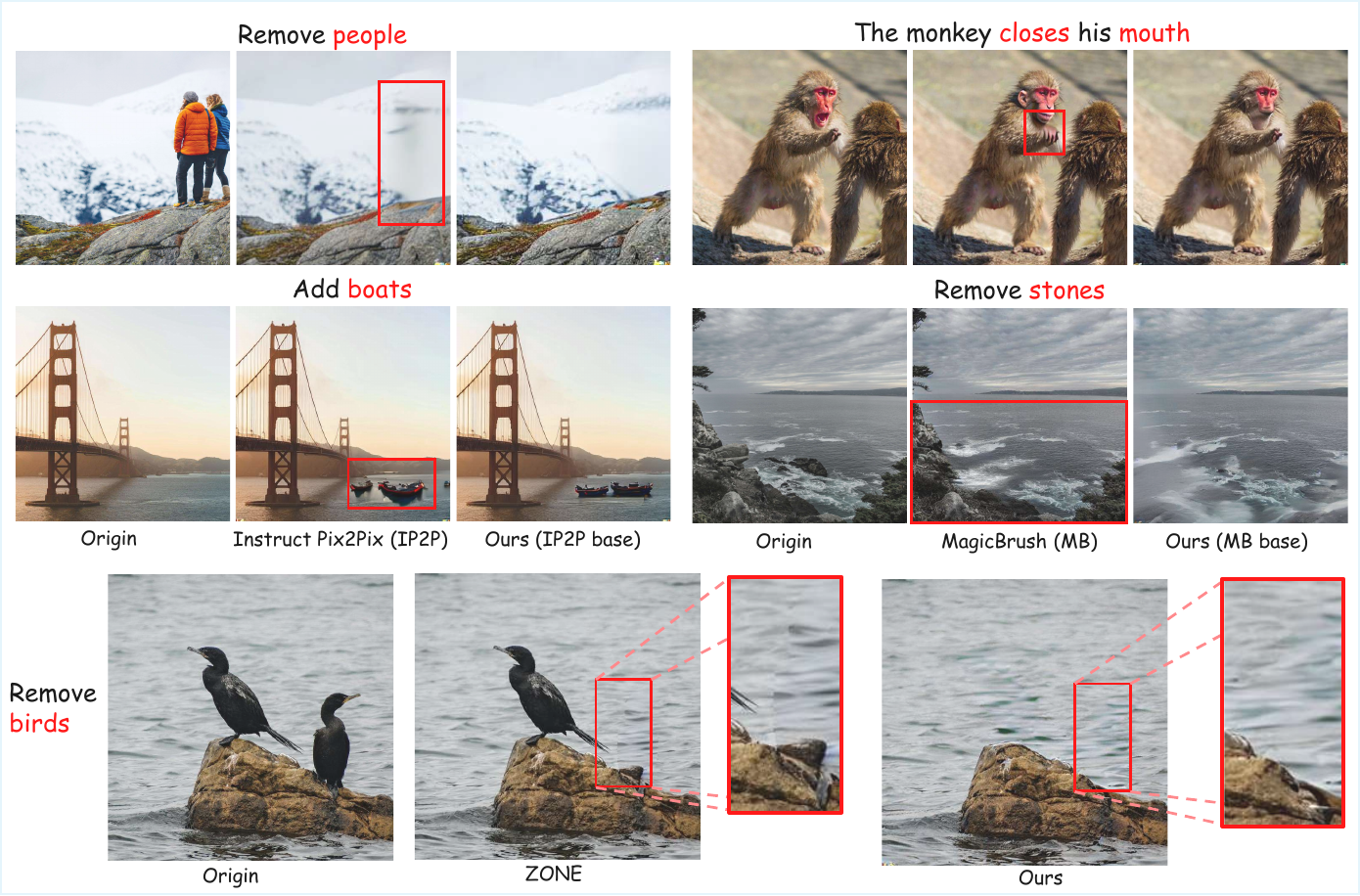}
    \caption{Qualitative comparison with baseline methods. Our framework produces edits that are both semantically accurate and visually coherent across various instructions. Our results consistently exhibit better blending and alignment with both regional and global semantics.}
    \label{fig:qualitative_comparison}
    \vspace{-0.8cm}
\end{figure*}
As shown in the Table \ref{tab:quantitative_eval}, our framework consistently enhances the performance of existing diffusion-based editing models across all evaluation metrics. Applied to InstructPix2Pix, we observe a +21.5\% increase in CLIP-I (from 0.7531 to 0.9146) and a 59.2\% reduction in LPIPS (from 0.5008 to 0.2042), alongside a substantial FID drop from 139.60 to 77.07. Similar improvements are seen with MagicBrush, where CLIP-T improves from 0.9057 to 0.9207, and LPIPS decreases from 0.2448 to 0.2263. When integrated into ZONE, our method achieves state-of-the-art results, with CLIP-I of 0.9729, DINO of 0.9955, and FID reduced by 38.3\% (from 44.63 to 27.53). These consistent gains across semantic, perceptual, and generative metrics highlight the generalizability and effectiveness of our framework in improving both local edit fidelity and global scene coherence.\\

As illustrated in Fig.~\ref{fig:qualitative_comparison}, our framework improves the visual quality and semantic consistency of edited images across various scenarios. For the prompt \textit{``remove people''}, InstructPix2Pix (IP2P) manages to erase the figures but leaves behind artifacts and unnatural textures in the edited area, whereas our method produces a more seamless and visually coherent background. Similarly, in the \textit{``add boats''} case, IP2P fails to integrate the boats naturally into the scene. With MagicBrush (MB), when prompted to \textit{``The monkey closes his mouth''}, the baseline not only alters the mouth but also distorts unrelated regions such as the hand, resulting in an unnatural appearance. Moreover, in the \textit{``remove stones''} example, MB cannot nearly remove these objects. Although ZONE exhibits strong localization due to its segmentation-based approach, the edited regions often appear visually detached from the rest of the image. This is evident in cases like \textit{``remove birds''}, where the edit is structurally correct but lacks smooth blending with the background. In comparison, our framework maintains both spatial accuracy and perceptual coherence, yielding edits that are semantically faithful and visually harmonious.

\section{Ablation studies}
\begin{figure}[h]
    \centering
    \begin{minipage}[h]{0.58\linewidth}
        \captionof{table}{Effectiveness of the proposed training framework through comparing models finetuned with and without our framework.}
        \label{tab:ablation1}
        \setlength{\tabcolsep}{5pt}
        \renewcommand{\arraystretch}{1.2}
        \begin{tabular}{l|ll|ll}
        \hline
        \hline
        \textbf{Model} & \textbf{\begin{tabular}[c]{@{}l@{}}IP2P \\ (without)\end{tabular}} & \textbf{\begin{tabular}[c]{@{}l@{}}IP2P\\ (with)\end{tabular}} & \textbf{\begin{tabular}[c]{@{}l@{}}MB\\ (without)\end{tabular}} & \textbf{\begin{tabular}[c]{@{}l@{}}MB\\ (with)\end{tabular}} \\ \hline
        CLIP-I↑ & 0.9054 & \textbf{0.9146} & 0.8793 & \textbf{0.9040} \\
        DINO↑ & 0.9680 & \textbf{0.9736} & 0.9517 & \textbf{0.9664} \\
        CLIP-T↑ & 0.9249 & \textbf{0.9337} & 0.9046 & \textbf{0.9207} \\
        LPIPS↓ & 0.2287 & \textbf{0.2042} & 0.2696 & \textbf{0.2263} \\
        FID↓ & 83.13 & \textbf{77.07} & 104.82 & \textbf{85.61} \\
        IS↑ & 9.6015 & \textbf{10.4931} & 8.9593 & \textbf{9.6078} \\ 
        \hline
        \hline
        \end{tabular}
    \end{minipage}%
    \hfill
    \begin{minipage}[h]{0.4\linewidth}
        \centering
        \includegraphics[width=0.9\linewidth]{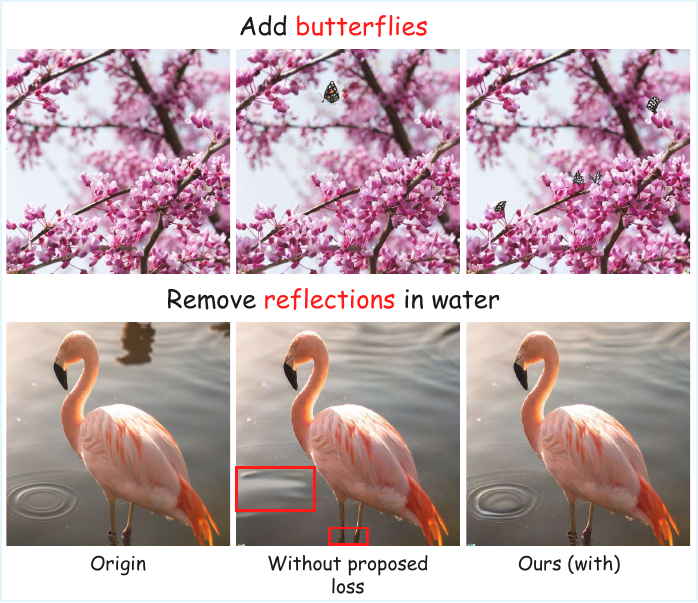}
        \caption{Qualitative ablation study on loss components: the model with vs without proposed loss.}
        \label{fig:ablationnoproposed}
    \end{minipage}
\end{figure}

We have finetuned the models IP2P~\cite{ip2p} and MagicBrush (MB)~\cite{zhang2024magicbrushmanuallyannotateddataset} without our proposed framework, using the identical experimental setup described in Sec.~\ref{implement}, to demonstrate that the observed performance gains stem from our training framework rather than from other experimental factors. The results, shown in Table~\ref{tab:ablation1}, clearly indicate consistent improvements across all metrics when our framework is applied. Specifically, for IP2P, our method improves CLIP-I from 0.9054 to 0.9146 and reduces LPIPS from 0.2287 to 0.2042, demonstrating better semantic preservation and perceptual fidelity. FID also drops from 83.13 to 77.07, with IS increasing from 9.6015 to 10.4931. Similarly, for MB, CLIP-T improves from 0.9046 to 0.9207, and FID decreases significantly from 104.82 to 85.61. These results confirm that our framework generalizes well across architectures and yields better alignment between the generated image and both regional and global textual guidance.
Qualitatively, as illustrated in Fig.~\ref{fig:ablationnoproposed}, we observe that models trained without our proposed framework often perform imprecise or overly broad edits. For instance, with the prompt \textit{``Remove reflection in water''}, the model trained without our method not only removes the reflection but also unintentionally alters unrelated regions such as the water surface and the bird's feet. In contrast, our approach yields more focused and semantically faithful edits, preserving surrounding details while accurately fulfilling the instruction. \\
\vspace{-0.3cm}
\begin{table}[h]
\centering
\caption{Evaluate the impact of removing each component in our training framework. Region-level semantic alignment and gated fusion contribute significantly to the model’s performance, as removing either leads to notable degradation in quality.}
\label{tab:ablation2}
\setlength{\tabcolsep}{5pt}
\renewcommand{\arraystretch}{1.2}
\begin{tabular}{lcccc}
\hline
\hline
\textbf{Model} & \textbf{\begin{tabular}[c]{@{}c@{}}IP2P \\ (without \\ full desc loss)\end{tabular}} & \textbf{\begin{tabular}[c]{@{}c@{}}IP2P \\ (without \\region desc loss)\end{tabular}} & \textbf{\begin{tabular}[c]{@{}c@{}}IP2P \\ (without \\ gated fusion)\end{tabular}} & \textbf{\begin{tabular}[c]{@{}c@{}}IP2P \\ (with \\ proposed loss)\end{tabular}} \\ \hline
CLIP-I↑ & 0.8851 & 0.8806 & 0.8688 & \textbf{0.9146} \\
DINO↑ & 0.9565 & 0.9569 & 0.9518 & \textbf{0.9736} \\
CLIP-T↑ & 0.9089 & 0.9136 & 0.8996 & \textbf{0.9337} \\
LPIPS↓ & 0.2583 & 0.2600 & 0.2942 & \textbf{0.2042} \\
FID↓ & 102.63 & 99.59 & 102.79 & \textbf{77.07} \\
IS↑ & 9.5514 & 9.1468 & 9.2767 & \textbf{10.4931} \\ 
\hline
\hline
\vspace{-1.15cm}
\end{tabular}
\end{table}
\begin{figure}[h]
    \centering
    \begin{minipage}{0.5\linewidth}
        \centering
        \includegraphics[width=\linewidth]{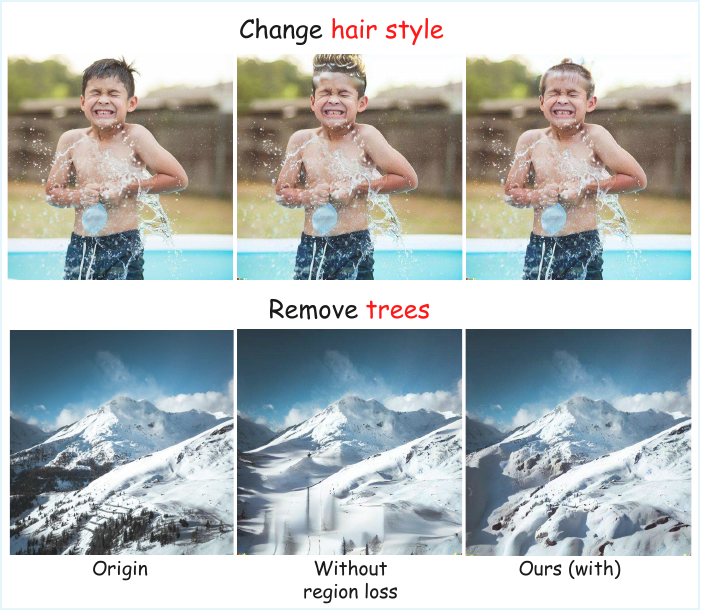}
        \caption*{(a) Without region-level loss} 
    \end{minipage}%
    \hfill
    \begin{minipage}{0.5\linewidth}
        \centering
        \includegraphics[width=\linewidth]{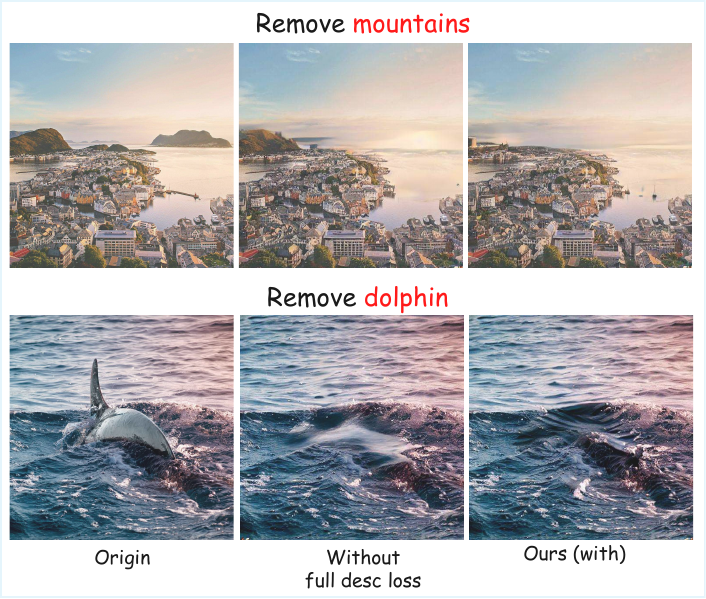}
        \caption*{(b) Without full-image description loss}
    \end{minipage}
    
    \caption{Visual comparison of models trained without region-level (left) and full-image (right) description losses.}
    \label{fig:ablation_combined}
\end{figure}
\vspace{0.5cm}
To better understand the contribution of each component in our framework, we conduct an ablation study by selectively removing key elements: the full description loss, the region description loss, and the gated fusion module. As shown in Table~\ref{tab:ablation2}, removing the region description loss results in greater performance degradation (e.g., CLIP-I drops from 0.9146 to 0.8806 and FID increases from 77.07 to 99.59) than removing the full description loss (CLIP-I drops to 0.8851, FID increases to 102.63), indicating that localized semantic alignment plays a more critical role in image editing. Moreover, disabling the gated fusion while keeping the region description loss leads to the worst performance across most metrics (e.g., LPIPS rises to 0.2942 and CLIP-I drops to 0.8688), emphasizing the importance of integrating contextual information during region embedding. These findings support our hypothesis that both region-aware guidance and context fusion are essential for achieving fine-grained, coherent edits.

Fig.~\ref{fig:ablation_combined} illustrates the qualitative improvements achieved by incorporating our proposed loss components. For instance, in the example of \textit{``remove dolphin''}, although the baseline model (trained without our losses) succeeds in removing the dolphin, it leaves behind an unnaturally flat water region. In contrast, our method produces a more visually coherent result by synthesizing natural water textures—such as waves—within the edited region, blending it seamlessly into the context. 
\section{Conclusion}
In this work, we propose a novel region-aware training framework for text-conditioned image editing that integrates both local and global semantic alignment. Inspired by how humans contextualize edits within an entire scene, our method enables each editable region to understand its role in the broader visual composition through a gated region-context fusion module and dual-level vision-language supervision. We optimize region-level embeddings using contrastive loss against fine-grained textual descriptions and align the entire image with detailed scene-level descriptions generated by a vision-language model. Extensive experiments demonstrate that our framework significantly improves edit accuracy, semantic coherence, and visual fidelity across multiple pretrained diffusion-based editors. These results confirm the effectiveness and generalizability of our approach in achieving fine-grained, contextually consistent image editing.
{

}
\newpage
\section*{Appendix}
\begin{table*}[h!]
\caption{Prompt for the LLM to make the image description in a template}
\renewcommand{\arraystretch}{1.2}
\begin{tabularx}{\textwidth}{l|X}
\hline
\hline
\textbf{Prompt} & 
\begin{minipage}[t]{\linewidth}
\begin{itemize}[left=0pt, itemsep=4pt, topsep=2pt]
    \item \textbf{Input:} \texttt{<image>}
    \item \textbf{Instruction:} You are a meticulous visual analyst. Carefully examine the given image and describe it in a single, flowing paragraph (maximum 520 tokens). Focus on every visually observable detail—such as color, texture, material, size, shape, and spatial relationships. Do not use bullet points or lists.
    
    \item \textbf{Constraints:} Avoid assumptions or inferences about unseen factors (e.g., time of day, season, emotions, story). Describe only what is directly visible in the image.

    \item \textbf{Your paragraph must naturally include:}
    \begin{itemize}
        \item A clear overview of the setting (e.g., indoor/outdoor, environment type, lighting conditions, background elements, overall mood)
        \item Detailed description of each major object:
        \begin{itemize}
            \item Appearance, color, material (wood, metal, fabric, etc.)
            \item Texture (smooth, rough, shiny, soft, etc.)
            \item Size (relative to others)
            \item Spatial position (e.g., foreground, center-left)
        \end{itemize}
        \item If humans or animals are present, describe each one in full detail:
        \begin{itemize}
            \item Hair, face, visible skin or fur, and accessories
            \item Clothing: color, texture, material, style, condition
            \item Pose: orientation and position of each body part (head, arms, legs, torso, hands, feet)
            \item Stance or motion—only if clearly visible and grounded in the image
        \end{itemize}
        \item Distinct, thorough description for multiple people or animals
        \item Supporting/background elements: furniture, walls, ground, vegetation, distant objects
        \item Clear spatial relationships (e.g., in front of, behind, next to, overlapping, under)
        \item Explicit description of the visual features of each object or region targeted in the editing instruction: \texttt{"\{edit\_prompt\}"}. For example, if the instruction is \texttt{"The girl bent and raised her two hands"}, then describe:
        \begin{itemize}
            \item Her posture (e.g., leaning forward, bent knees)
            \item The position and gesture of her hands (e.g., raised above shoulders, palms open)
        \end{itemize}
    \end{itemize}

    \item \textbf{Style Requirement:} Use vivid, sensory-rich language. Every detail must be grounded in what can actually be seen. Avoid summarizing—immerse the reader in a scene constructed entirely from the image's visible content.\\
    
\end{itemize}
\end{minipage}
\\
\hline
\hline
\end{tabularx}
\label{tab:llmprompt}
\end{table*}

\begin{table*}[ht]
\centering
\caption{Some examples of the image description generated by Deepseek-VL-Tiny with the prompt in Table \ref{tab:llmprompt}}
\renewcommand{\arraystretch}{1.5}
\begin{tabular}{>{\centering\arraybackslash}m{0.25\linewidth} | >{\arraybackslash}m{0.13\linewidth} | >{\arraybackslash}m{0.53\linewidth}}
\toprule
\toprule
\textbf{Image} & \textbf{Editing Instruction} & \textbf{Image Description} \\
\midrule

\includegraphics[width=0.9\linewidth]{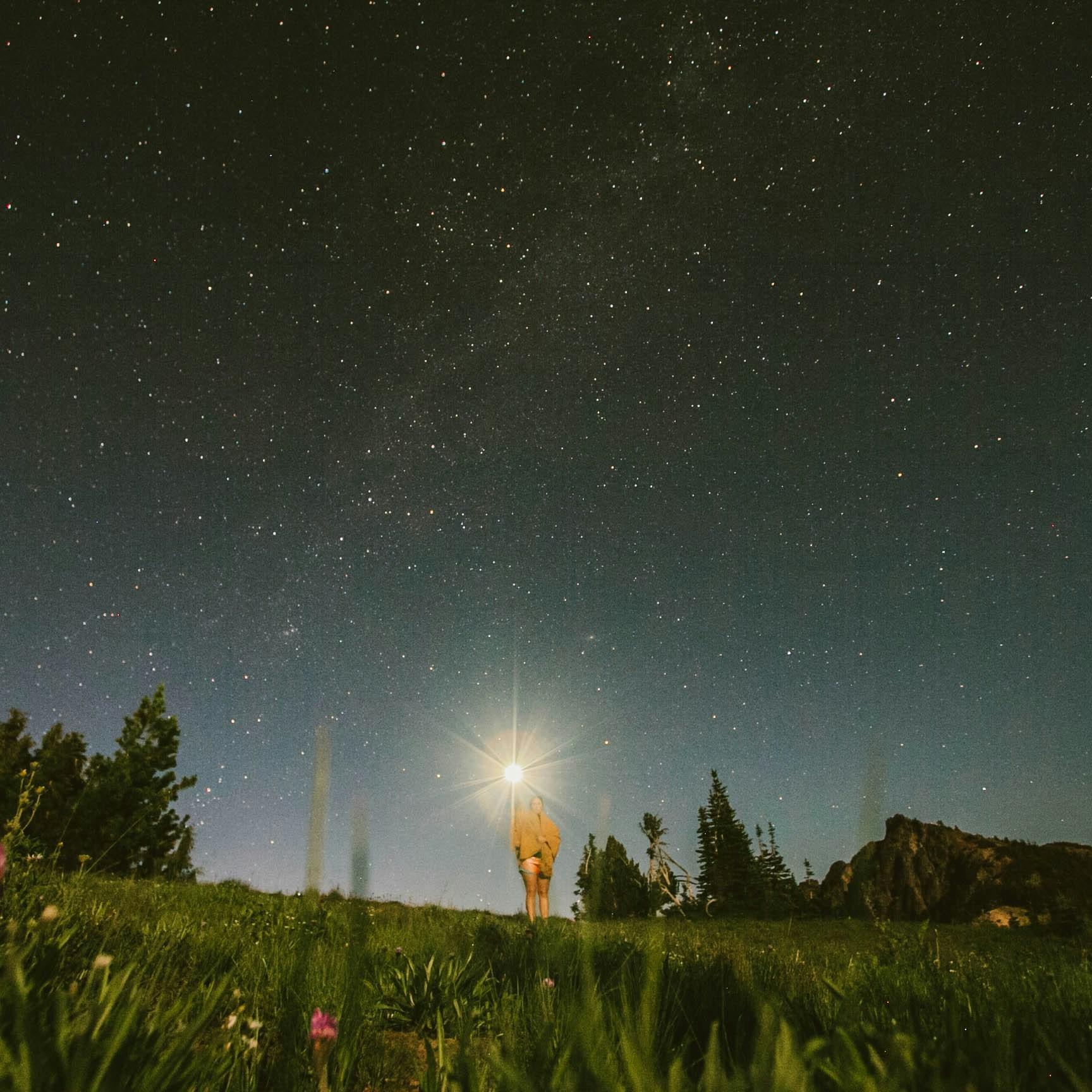} 
& Remove the woman. 
& The image captures a serene nighttime landscape, where the starry sky dominates the upper portion of the frame. The celestial display is dense with stars, creating a textured canvas of light and dark. Below, the terrain is a lush green field, dotted with wildflowers and grasses. In the middle ground, a cluster of coniferous trees stands tall, their silhouettes contrasting against the bright moon. The moon itself is a focal point, casting a soft glow that illuminates the scene with a gentle light. To the right, a rocky outcrop rises, adding a rugged element to the otherwise tranquil setting. The overall mood is peaceful and contemplative, inviting the viewer to pause and appreciate the beauty of the night sky. \\
\midrule

\includegraphics[width=0.9\linewidth]{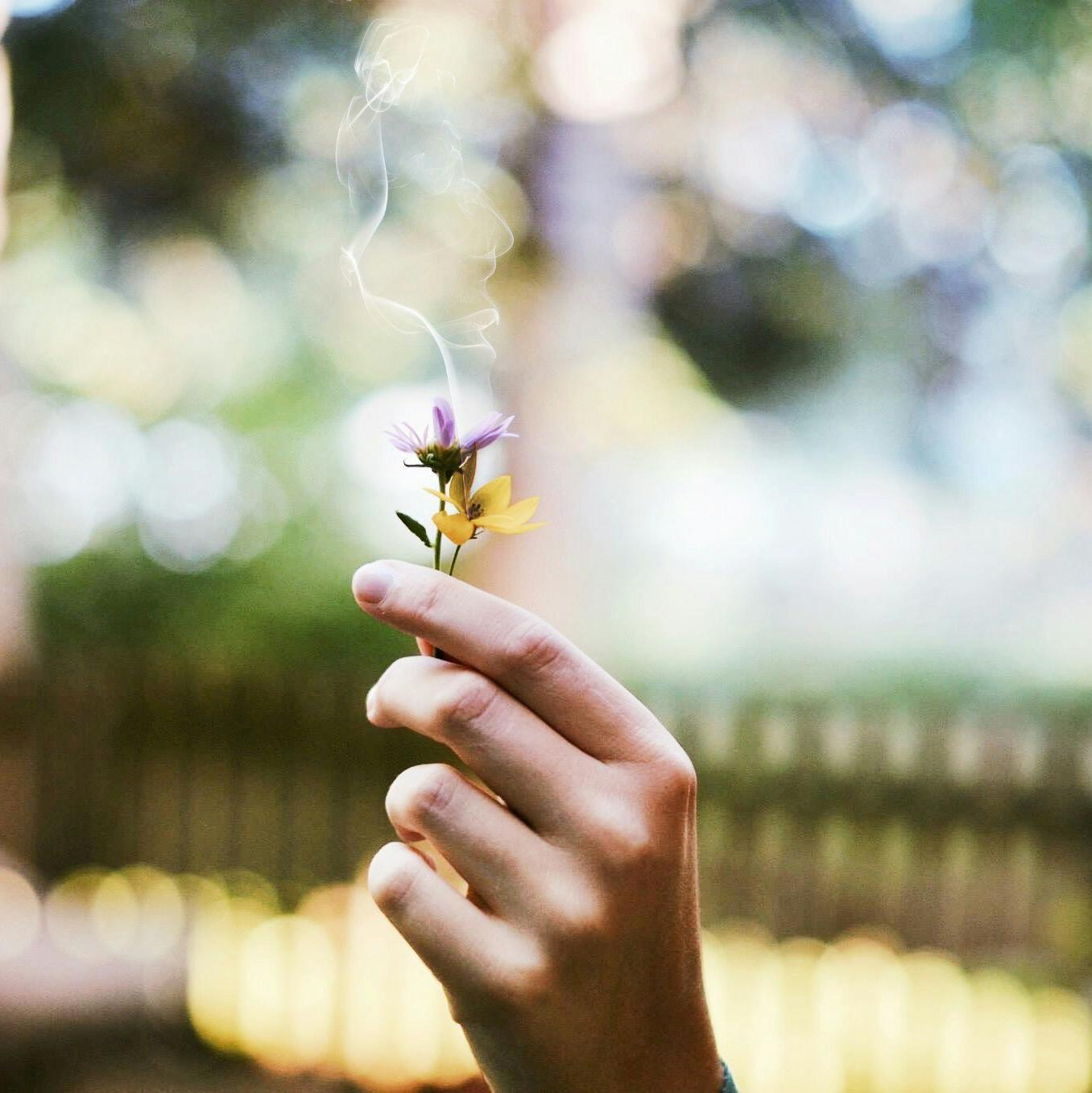} 
& Replace wildflowers with red roses. 
& The image captures a close-up view of a hand holding a single red rose. The rose is vibrant, with deep red petals that appear soft and velvety, suggesting it might be made of silk or satin. The stem is slender and green, indicating a natural material. The hand is positioned in the foreground, with fingers gently curled around the stem, showcasing the delicate texture of the skin. The background is blurred, creating a bokeh effect that highlights the rose and hand, giving the image a dreamy, ethereal quality. The lighting is soft and diffused, casting gentle shadows and enhancing the rich color of the rose. The overall mood is intimate and serene, evoking a sense of romance and tranquility. \\
\midrule

\includegraphics[width=0.9\linewidth]{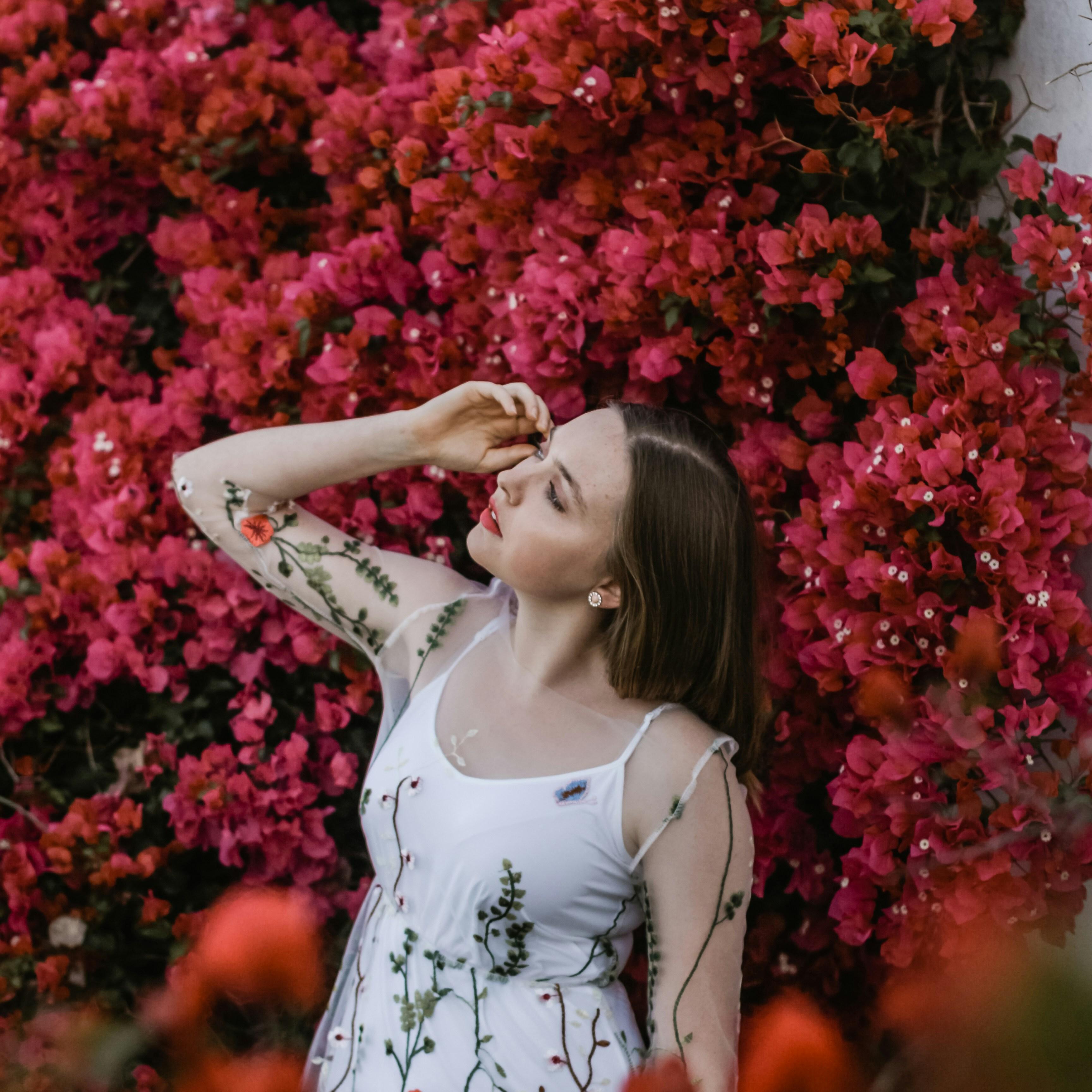} 
& Change a white dress into a black dress. 
& The image depicts a young woman standing amidst a vibrant backdrop of pink bougainvillea flowers. She is wearing a black dress adorned with white floral patterns, which contrasts beautifully with the bright colors of the flowers. Her hair is styled in a sleek, straight manner, and she has a subtle smile on her face. The lighting is soft and natural, suggesting that the photo was taken outdoors during the daytime. The overall mood of the image is serene and elegant, with the woman appearing relaxed and at ease in her surroundings. \\
\midrule

\includegraphics[width=0.9\linewidth]{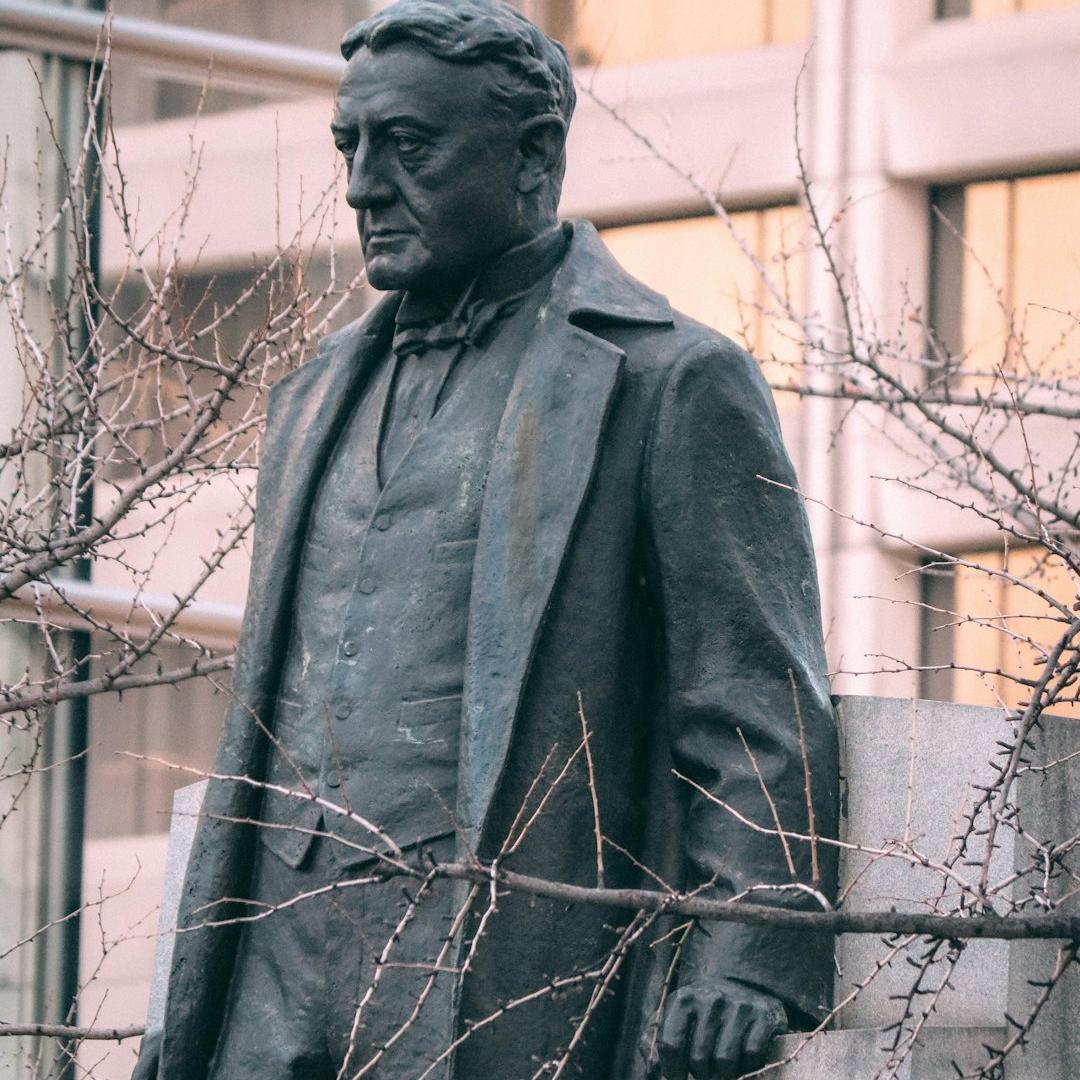} 
& Add a bird 
& The image depicts a bronze statue of a man in formal attire, including a suit jacket, vest, and bow tie. The statue is situated outdoors, with bare branches of a tree partially obscuring the view. A bird is perched on the shoulder of the statue, adding a touch of life to the scene. The background features a modern building with large windows, suggesting an urban environment. The lighting appears soft, possibly indicating an overcast day or the time of day being late afternoon. The statue's detailed craftsmanship is evident in the texture of the clothing and the realistic portrayal of the man's features. The bird adds a dynamic element to the otherwise static composition. \\
\bottomrule
\bottomrule
\end{tabular}
\label{tab:edit_table}
\end{table*}

\begin{figure*}[h]
    \centering
    \includegraphics[width=\linewidth]{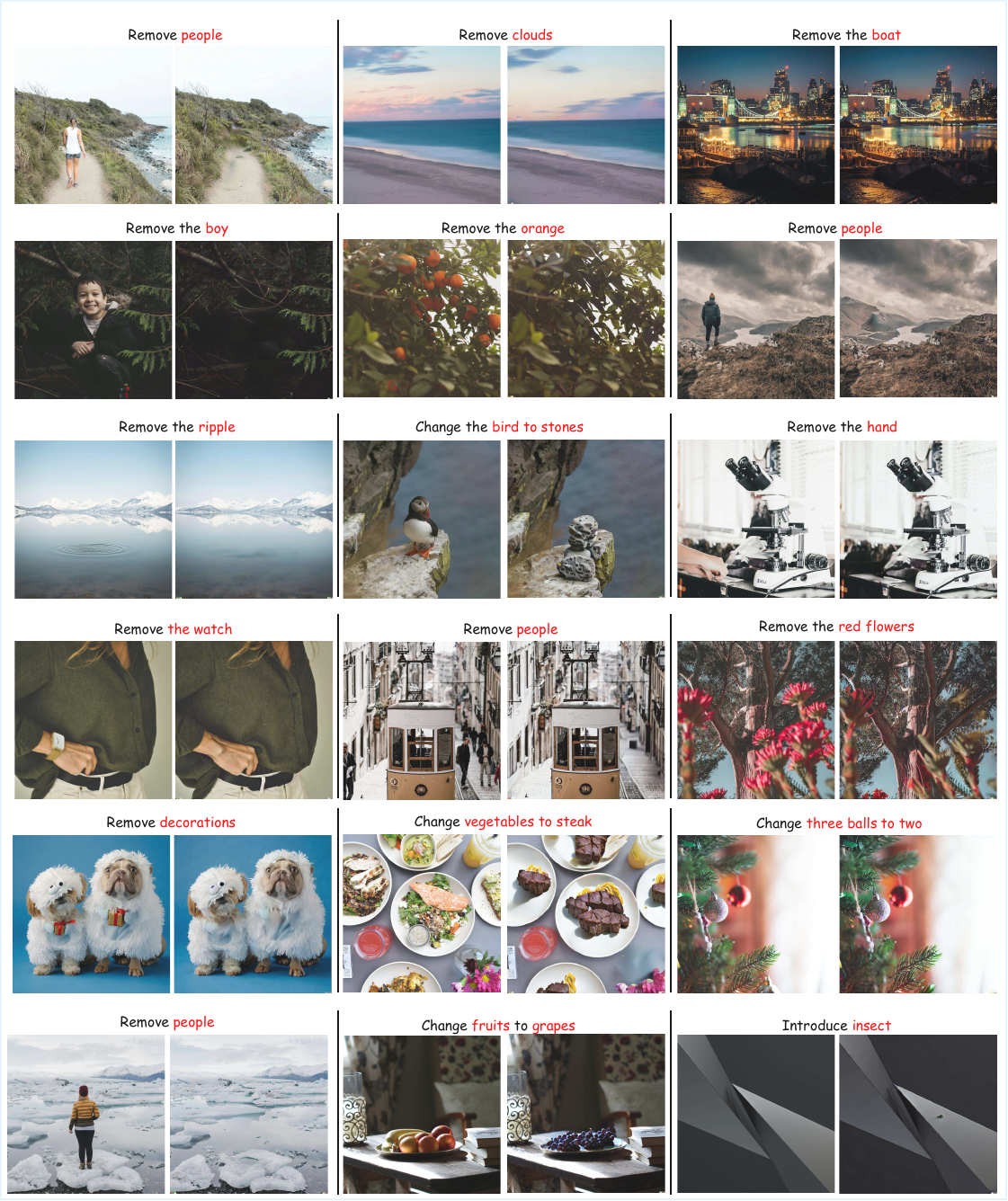}
    \caption{Some qualitative results of our proposal}
    \label{fig:qualitative_performance}
\end{figure*}
\end{document}